\documentclass[letterpaper]{article} 
\usepackage{aaai25}  
\usepackage{times}  
\usepackage{helvet}  
\usepackage{courier}  
\usepackage[hyphens]{url}  
\usepackage{graphicx} 
\usepackage{natbib}  
\usepackage{caption} 
\usepackage{algorithm}
\usepackage{algorithmic}
\usepackage{multirow} 
\usepackage[cmyk]{xcolor}
\usepackage{bbding}
\usepackage{utfsym}
\usepackage{subcaption} 
\usepackage{graphicx}
\definecolor{LightGray}{gray}{0.9} 
 
\usepackage{newfloat}
\usepackage{listings}
\usepackage{booktabs} 
\usepackage{amsmath}
\usepackage{amssymb}
\DeclareCaptionStyle{ruled}{labelfont=normalfont,labelsep=colon,strut=off} 
\floatstyle{ruled}
\newfloat{listing}{tb}{lst}{}
\floatname{listing}{Listing}
\title{Enriching Multimodal Sentiment Analysis through Textual Emotional Descriptions of Visual-Audio Content}
\author{
    Sheng Wu\textsuperscript{\rm 1,2}, 
    Xiaobao Wang\textsuperscript{\rm 3,2,}\equalcontrib,
    Longbiao Wang\textsuperscript{\rm 3},
    Dongxiao He\textsuperscript{\rm 3},
    Jianwu Dang\textsuperscript{\rm 4}
}
\affiliations{
    \textsuperscript{\rm 1}School of New Media and Communication, Tianjin University, Tianjin, China\\

    \textsuperscript{\rm 2}Guangdong Laboratory of Artificial Intelligence and Digital Economy (SZ)\\
    \textsuperscript{\rm 3}Tianjin Key Laboratory of Cognitive Computing and Application, College of Intelligence and Computing, Tianjin University, Tianjin, China\\
    \textsuperscript{\rm 4}Shenzhen Institute of Advanced Technology, Chinese Academy of Sciences, Shenzhen, China\\
    \{2022245014, wangxiaobao, longbiao\_wang, hedongxiao\}@tju.edu.cn, jdang@jaist.ac.jp

}

\usepackage{bibentry}

\begin{document}

\maketitle

\begin{abstract}
Multimodal Sentiment Analysis (MSA) stands as a critical research frontier, seeking to comprehensively unravel human emotions by amalgamating text, audio, and visual data. Yet, discerning subtle emotional nuances within audio and video expressions poses a formidable challenge, particularly when emotional polarities across various segments appear similar.
In this paper, our objective is to spotlight emotion-relevant attributes of audio and visual modalities to facilitate multimodal fusion in the context of nuanced emotional shifts in visual-audio scenarios. To this end, we introduce DEVA, a progressive fusion framework founded on textual sentiment descriptions aimed at accentuating emotional features of visual-audio content. DEVA employs an Emotional Description Generator (EDG) to transmute raw audio and visual data into textualized sentiment descriptions, thereby amplifying their emotional characteristics. These descriptions are then integrated with the source data to yield richer, enhanced features.
Furthermore, DEVA incorporates the Text-guided Progressive Fusion Module (TPF), leveraging varying levels of text as a core modality guide. This module progressively fuses visual-audio minor modalities to alleviate disparities between text and visual-audio modalities.
Experimental results on widely used sentiment analysis benchmark datasets, including MOSI, MOSEI, and CH-SIMS, underscore significant enhancements compared to state-of-the-art models. Moreover, fine-grained emotion experiments corroborate the robust sensitivity of DEVA to subtle emotional variations.
\end{abstract}

%

\section{Introduction}

\begin{figure}[htbp]
  \centering
  \includegraphics[width=\linewidth]{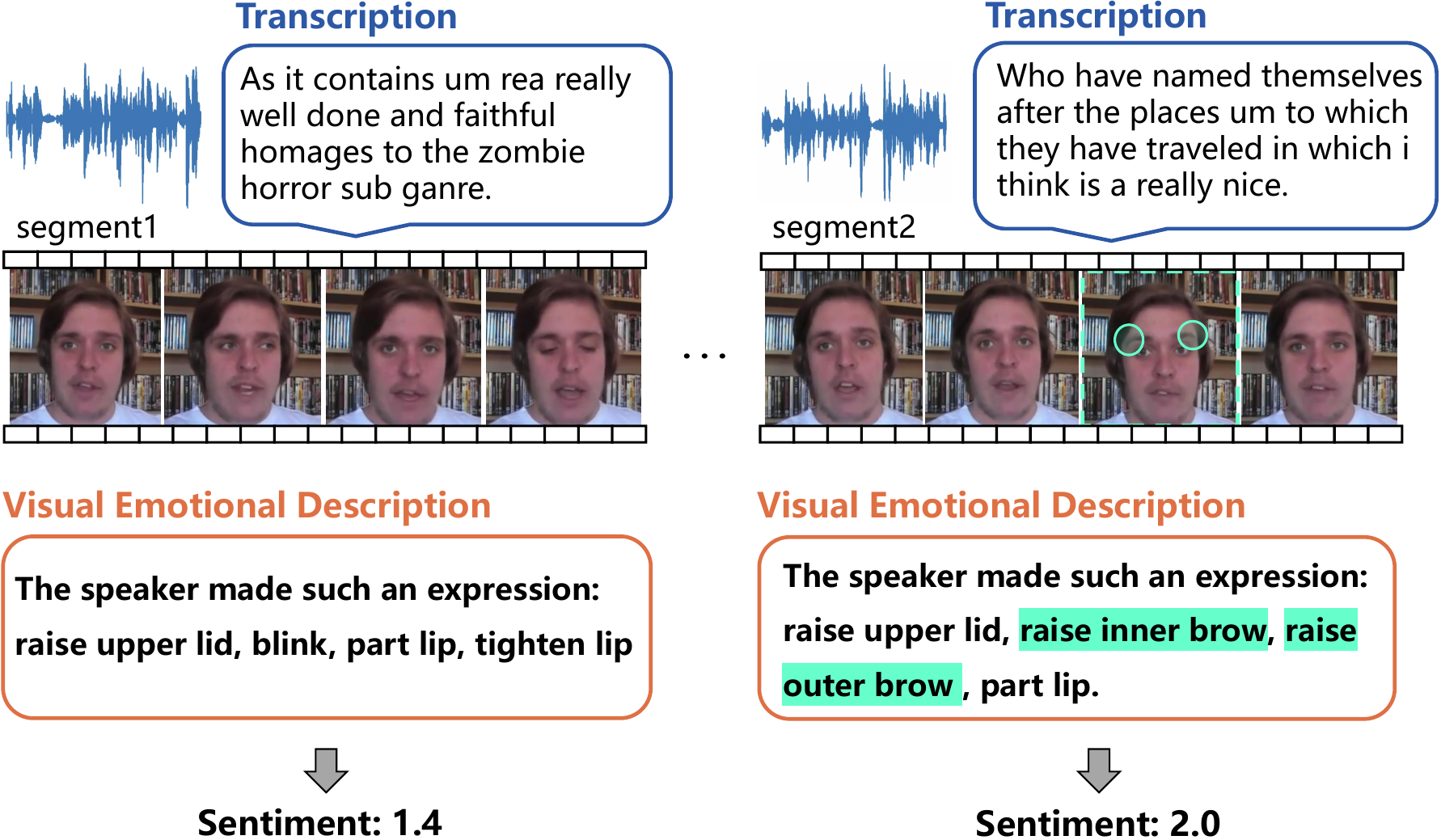}
  \caption{The illustration of our motivation is as follows: The text transcription of the audio is indicated within the blue box, while the visual emotional description is within the orange box. The teal highlighting indicates highly emotionally relevant descriptions, and teal circles are used to mark the corresponding microexpressions in the source data.}
  \label{fig:intro}
\end{figure} 

Sentiment analysis is a classic language understanding task, where traditionally, the analysis of user sentiment is conducted through text ~\cite{Lei2016RatingPB}. With the evolution of social media, there has been a significant increase in user-generated videos, making multimodal sentiment analysis (MSA) gradually emerge as a hotspot in research. Its objective is to comprehensively analyze people's emotional states through text, audio, and visual data~\cite{Poria2020BeneathTT, Dong2024UnveilingID}. MSA plays a crucial role in various fields such as healthcare~\cite{Doctor2016AnIF,Jiang2020ASR}, social media analytics~\cite{Melville2009SentimentAO}, and human-computer interaction~\cite{Peter2012EmotionIH}. Compared to unimodal approaches, multimodal analysis offers greater robustness and comprehensiveness in understanding human emotions.

Leveraging the advancements in deep learning techniques~\cite{Chen2022ViT,Huang2021CMUAWatermarkAC,Fu2021CONSKGCNCS, Wang2023AugmentingAD}, recent approaches in multimodal sentiment analysis (MSA) primarily concentrate on representation learning and fusion strategies across modalities. In terms of representation learning, various methods have emerged, including feature decoupling techniques aimed at mapping features into shared and private spaces~\cite{Hazarika2020MISA, Yang2022FDMER, Yang2022MFSA}. Moreover, contrastive learning~\cite{Yang2023ConFEDE} and multitask learning~\cite{Yu2021SelfMM} have been explored to enhance representation learning. Regarding multimodal fusion, the initial approach is often early fusion, where features from text, audio, and visual modalities are concatenated for downstream tasks. Subsequently, more sophisticated techniques such as outer product~\cite{Zadeh2017TFN}, Convolutional Neural Networks~\cite{Huang2020MultimodalER}, and Recurrent Neural Networks~\cite{Sun2020MultimodalCD} have been adopted. Furthermore, attention mechanisms~\cite{Chen2017MultimodalSA,Tsai2019MulT} have been explored for multimodal data fusion in recent studies.

However, we find that discerning subtle differences in emotional intensity becomes challenging when the emotional polarities of different segments are closely aligned, particularly when analyzing raw data such as audio and visual inputs. 
As depicted in Figure \ref{fig:intro}, segments 1 and 2 demonstrate highly similar audio temporal distributions, and their overall facial expressions exhibit considerable resemblance. 
Relying solely on transcription, audio, and visuals poses difficulties in accurately determining the sentimental polarity of segment 2. 
To the best of our knowledge, previous studies have not explicitly tackled the scenario of fine-grained emotional changes in visual-audio content.

Fortunately, several studies suggest that different modalities contribute disparately to the Multimodal Sentiment Analysis (MSA) task, with text often serving as the core modality, audio and visual as auxiliary modalities~\cite{Zhang2023ALMT,Wu2021TCSP,Li2022AMOA}. 
We find that articulating emotions in audio and depicting expressions using text can inherently accentuate disparities in sentimental polarity, leading to more precise sentimental assessments. 
This observation holds true across multiple datasets, as illustrated in Figure \ref{fig:intro}. For instance, when we describe facial expressions in emotional terms, the micro-expression of “raising eyebrows" translates into “raise inner brow, raise outer brow" in textual form, thereby suggesting that segment 2 elicits a more positive sentiment compared to segment 1.

Based on the above observations, we propose DEVA, a novel approach designed to accentuate emotional expressions in visual-audio content through textual descriptions. DEVA constructs these descriptions by narrating minor modalities with text and progressively integrating them with the source data under textual guidance. Initially, DEVA utilizes a pre-trained BERT model~\cite{Devlin2019BERTPO} alongside separate feature extractors for audio and visual modalities to process text, audio, and visual inputs. Following this, three Transformers encode each modality into a unified format. Subsequently, Emotional Description Generator (EDG) leverages OpenFace~\cite{Baltruaitis2018OpenFace2F} and OpenSMILE~\cite{Eyben2010OpensmileTM} tools to extract emotionally relevant visual and acoustic features from the video, generating natural language descriptions to highlight emotionally significant features, particularly subtle emotional shifts in audio and visual cues. These descriptions are then fused with the source data features to enhance modality features. Furthermore, we introduce the Text-Guided Progressive Fusion Module (TPF), utilizing text as the core modality to guide the fusion of audio and visual modalities into minor modality fusion features. Finally, the core and minor modality features are employed in cross-modal Transformers for fusion, effectively bridging distribution differences between the core and minor modalities.

The main contributions can be summarized as follows:
\begin{itemize}
\item We propose DEVA, a progressive fusion framework based on emotional description to highlight the emotional characteristics of visual-audio content. This method transforms audio and visual source data into textual emotional descriptions.
\item We introduce a novel Emotional Description Generator (EDG), which textualizes minor modalities into emotional descriptions to highlight the emotional representation in audio and visual, addressing the issue of fine-grained variations in visual-audio emotion features. Meanwhile, we design a Text-guided Progressive Fusion method (TPF), which progressively fuses audio and visual data into minor modality features guided by text to bridge the gap between core and minor modalities.
\item DEVA achieves state-of-the-art performance on three popular MSA datasets, thoroughly analyzing and validating the method's effectiveness and advancements.
\end{itemize}

\begin{figure*}[h]
  \centering
  \includegraphics[width=0.9\linewidth]{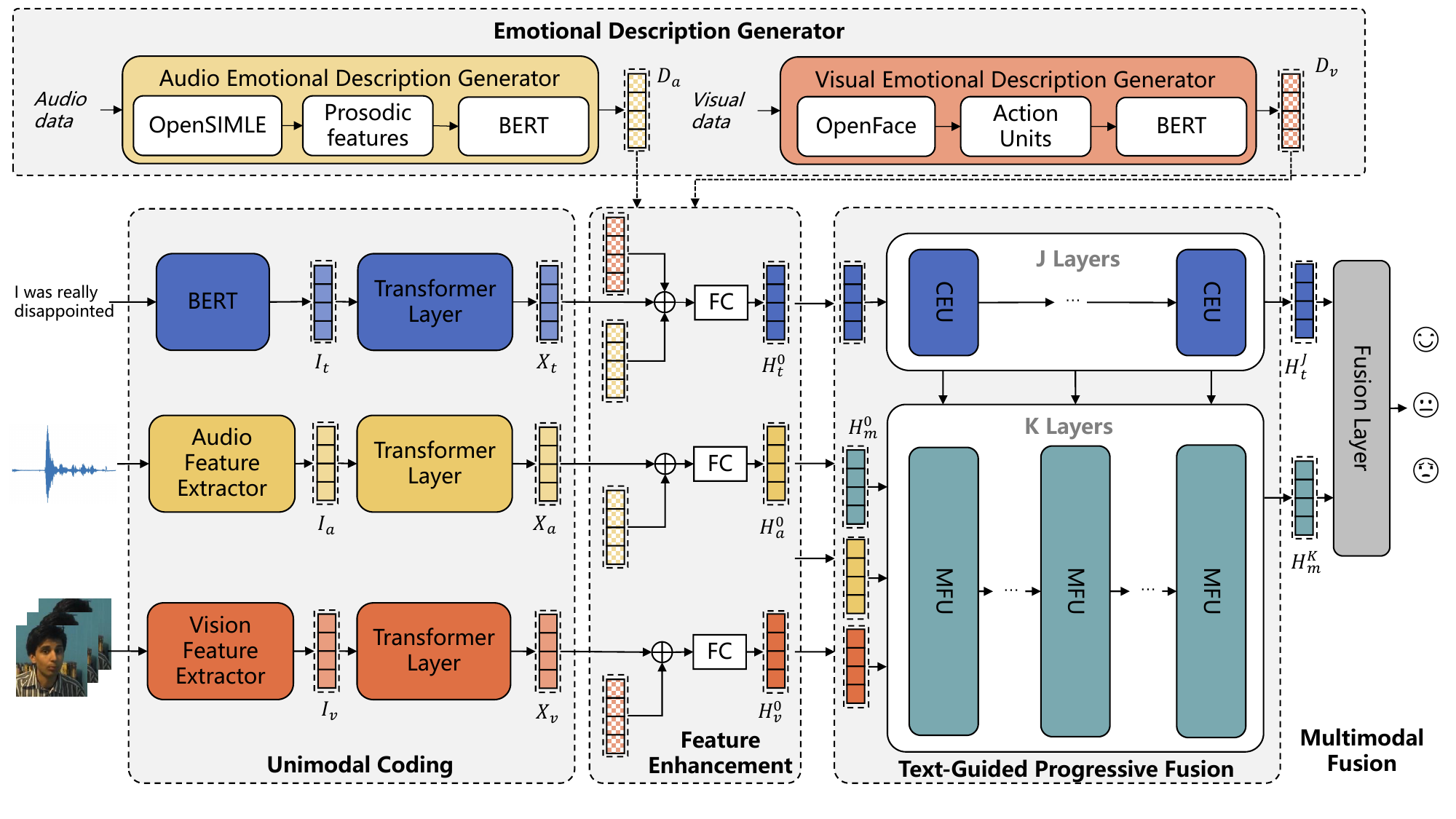}
  \caption{The overall architecture of DEVA consists of unimodal coding, an emotional description generator, feature enhancement, text-guided progressive fusion, and multimodal fusion.}
  \label{fig:method}
\end{figure*}

\section{Related Work}
Multimodal Sentiment Analysis (MSA) is a widely studied research topic. 
Unlike previous approaches that solely utilized unimodal data~\cite{Aman2007IdentifyingEO, Shirian2020CompactGA}, MSA integrates text, audio, and visual non-verbal information to obtain more rich and robust features.
Previous research has mainly focused on representation learning and multimodal fusion.
For methods centered around representation learning, approaches like~\cite{Hazarika2020MISA, Yang2022FDMER, Yang2022MFSA} decompose each modality into modality-invariant and -specific representations, utilizing squared norm loss as a constraint. 
In terms of methods focused on multimodal fusion, Zedeh ~\cite{Zadeh2016MultimodalSI} first proposed a multimodal dictionary, learning dynamic interactions between facial gestures and spoken vocabulary to model sentiment. 
Later, the Tensor Fusion Network was introduced~\cite{Zadeh2017TFN}, which encodes each modality with its corresponding sub-network and models interactions at the unimodal, bimodal, and trimodal levels through the triple Cartesian product. 
Approaches like~\cite{Zadeh2018MFN,Yu2021SelfMM,Yang2023ConFEDE} adopt late fusion at the decision level. 

In recent years, Transformers ~\cite{Vaswani2017AttentionIA} have dominated the field of deep learning. 
In the field of Multimodal Sentiment Analysis (MSA), Transformer is widely used for feature extraction, representation learning, and multimodal fusion.
The Tsai et al.~\cite{Tsai2019MulT} introduced a multimodal transformer to align multimodal sequences, but the pairwise fusion approach is less efficient.
~\cite{Lv2021ProgressiveMR,Zhang2023ALMT} combine textual and non-textual features for multimodal interaction and fusion. 
Unlike the aforementioned work, we partition each modality into traditional data and text data based on emotional descriptions, which complement each other, thereby obtaining a unimodal representation with enhanced expressive power.

\section{Method}

\subsection{Problem Definition}

The original inputs of multimodal sentiment analysis includes text ($t$), audio ($a$), and visual ($v$). 
The goal of this task is to fuse the data from different modalities and output the pretictive sentimental polarity $\hat{y}$.

\subsection{Model Overview}

The overall processing pipeline of the proposed Emotional Description Progressive framework (DEVA) is shown in Figure \ref{fig:method}. 
We begin by preprocessing the source data into a unified format. 
Simultaneously, the Emotional Description Generator (EDG) is employed to extract emotion-relevant features from audio and visual modality source data, transforming them into textual Emotional Description. 
These textual features are then concatenated with the source data and input into a fully connected layer for fusion enhancement. 
Subsequently, the Text-Guided Progressive Fusion module (TPF) is applied, using the core modality (text) from different Transformer layers as guidance to progressively fuse the minor modalities (audio and visual) and generate minor modality features. 
Finally, a cross-modal Transformer is utilized to merge core and minor modalities, yielding the ultimate multimodal representation.

\subsection{Unimodal Coding}
For unimodal encoding, we adopt a two-stage approach. In the first stage, following prior work, we utilize BERT~\cite{Devlin2019BERTPO} , Librosa~\cite{McFee2015librosaAA}, and OpenFace~\cite{Baltruaitis2018OpenFace2F} to individually encode text, audio, and visual source data, resulting in the representation $I_m \in \mathbb{R}^{T_m \times d_m}$, where $T_{m \in \{t,a,v\}}$ represents the length of each modal data, and $d_{m \in \{t,a,v\}}$ represents the dimension of each modal data.

In the second stage, we initialize a token $E_m$ for each modality and concatenate it after $I_m$, inputting the combined representation into the Transformer Layer~\cite{Vaswani2017AttentionIA} to obtain traditional features for each modality:
\begin{equation}
  X_m = Trans([I_m; E_m], {\theta}_{Trans}) \in \mathbb{R}^{T \times d},
\end{equation}
where $X_m$ is the unified feature of each modality $m \in \{t,a,v\}$ with a size of $T \times d$, $E_{Trans}$ and ${\theta}_{Trans}$ respectively represent the Transformer feature extractor and corresponding parameters, [$\cdot$;$\cdot$] represent the concatenation.

It is noteworthy that we do not use the output of the Transformer; instead, we select the first $T$ tokens (where $T < T_m$) as the traditional feature information. This choice is made because most of the information in the Transformer tends to aggregate in the embeddings of the initial tokens. Aggregating unimodal information in the initial tokens helps to eliminate redundant information.

\subsection{Emotional Description Generator}
We use a third-party tool to extract emotionally strongly correlated features from audio and visual (facial expressions) and convert them into text descriptions. Specifically, we describe the audio modality as text that contains loudness, pitch, jitter, and shimmer, and the visual modality with text that contains multiple facial expression action units.

\textbf{Audio Emotional Description.}
OpenSMILE~\cite{Eyben2010OpensmileTM} is a feature extractor used for audio signal processing, commonly utilized in fields such as speech recognition and affective computing. Through OpenSMILE, we extract four prosodic features closely related to emotion: pitch, loudness, jitter, and shimmer. Among them, jitter represents the variability in pitch, while shimmer represents the variability in loudness. We extract the prosodic features of the entire dataset and calculate the numerical tertiles for each of the four features, dividing them into low, normal, and high levels based on the tertiles. This results in $4 \times 3$ AED Units, as shown in the Table \ref{tab:PF}. Ultimately, we obtain AEDs in the following format: {\itshape "The Speaker made such an tone: pitch, loudness, jitter, and shimmer at different levels."}

\textbf{Visual Emotional Description.}
Since facial expression is the most important basis for visual judgment of sentiment, we utilize OpenFace~\cite{Baltruaitis2018OpenFace2F} to extract facial Action Units (AUs) as Visual Emotional Description. OpenFace provides 16 common AUs, each AU corresponding to a textual description. For instance, AU04 represents "lower brow," and AU20 represents "stretch lip." The specific AU descriptions are detailed in the Table \ref{tab:AU} below. Additionally, we establish a criterion for selecting AUs for description, which involves the following steps: a) If an AU appears continuously for three frames, it is considered a candidate. b)The candidate AUs are sorted based on their duration, from highest to lowest, and the top-$k$ AUs are selected for description. 

The reason for selecting $k$ AUs instead of all candidate AUs is to reduce the redundant impact of minor facial emotional features, allowing the model to capture the most significant emotional characteristics. Ultimately, we generate VEDs in the following format: {\itshape"The speaker made such an expression: $k$ AUs."}

Finally, we encode the obtained AED and VED using the same encoding method as the text modality to acquire emotional description $D_a \in \mathbb{R}^{T \times d}$ and $D_v \in \mathbb{R}^{T \times d}$:
\begin{equation}
    F^{'}_a = AEDG(I_a) \in \mathbb{R}^{T \times d},
\end{equation}
\begin{equation}
    F^{'}_v = VEDG(I_v) \in \mathbb{R}^{T \times d},
\end{equation}
\begin{equation}
  D_m = Trans([F^{'}_m; E_m]), {\theta}_{Trans}) \in \mathbb{R}^{T \times d},
\end{equation}
where $D_m$ is the unified feature of audio and visual emotional description, $m \in \{a,v\}$, and [$\cdot$;$\cdot$] represents the concatenation operation.

\begin{table}
  \resizebox{0.48\textwidth}{!}
  {
    \begin{tabular}{c c}
    \toprule
     Prosodic features & Description \\
    \midrule
    loudness & low loudness, normal loudness, high loudness \\
    pitch & low pitch, normal pitch, high pitch \\
    jitter & low jitter, normal jitter, high jitter \\
    shimmer & low shimmer, normal shimmer, high shimmer  \\
  \bottomrule
    \end{tabular}
    }

    \caption{Descriptions of Prosodic features.}
  \label{tab:PF}

\end{table}

\begin{table}
  \resizebox{0.48\textwidth}{!}
  {
  \begin{tabular}{c c c c}
    \toprule
     AU & Description & AU & Description \\
    \midrule
    AU01 & raise inner brow & AU12 & pull lip corner \\
    AU02 & raise outer brow & AU15 & depress lip corner \\
    AU04 & lower brow & AU20 & stretch lip \\
    AU05 & raise upper lid & AU23 & tighten lip  \\
    AU06 & raise cheek & AU25 & part lip \\
    AU07 & tighten lid & AU26 & drop jaw \\
    AU09 & wrinkle nose & AU28 & suck lip \\
    AU10 & raise upper lip & AU45 & blink \\
  \bottomrule
\end{tabular}
}

  \caption{Descriptions of Action Units.}
  \label{tab:AU}
\end{table}

\begin{figure}[h]
  \centering
  \includegraphics[width=\linewidth]{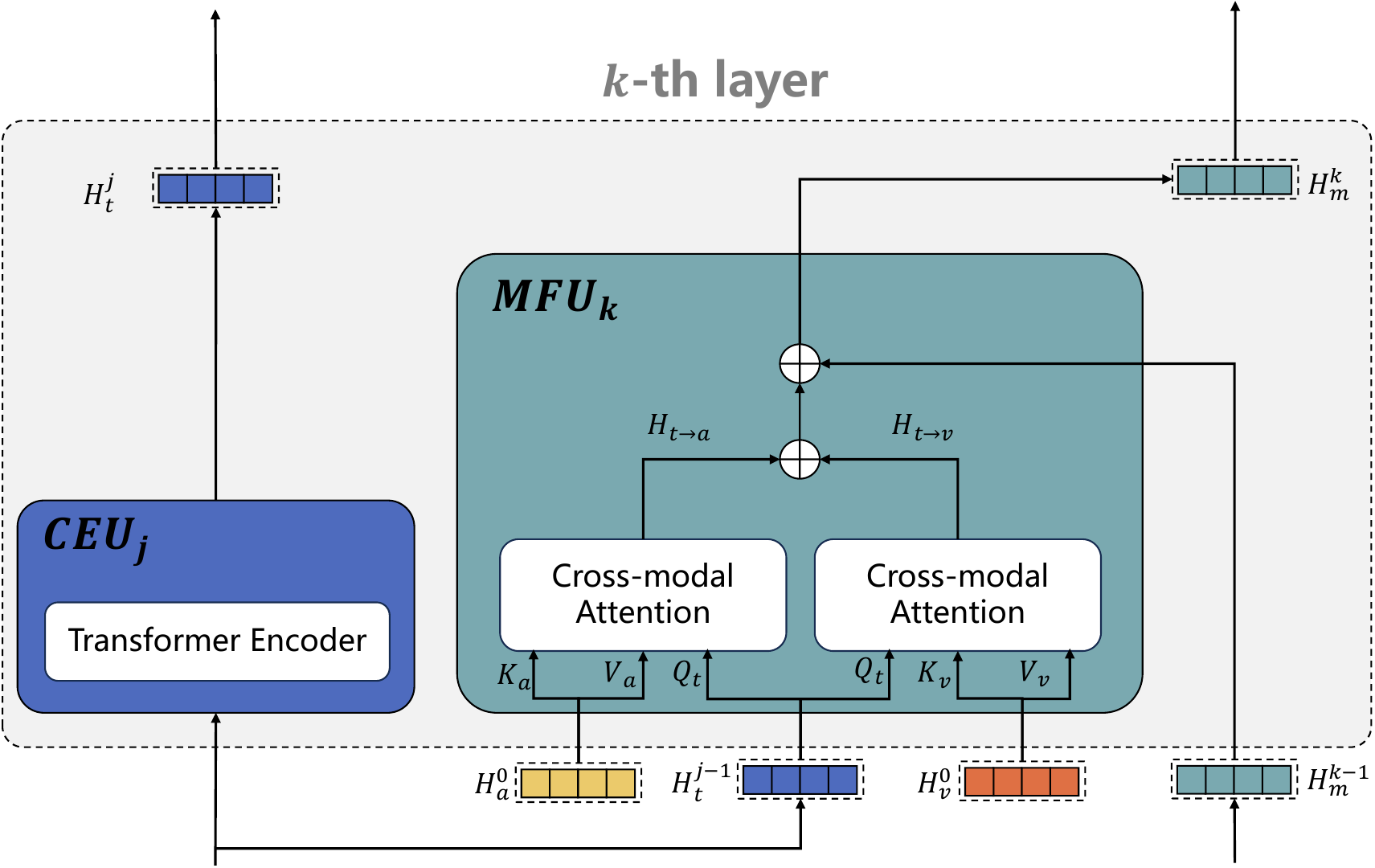}
  \caption{The architecture of TPF.}
  \label{fig:tpf}
\end{figure}

\subsection{Feature Enhancement}
We concatenate the traditional and description feature and merge them through fully connected layers, enhancing the traditional representation to include more emotional information. Specifically, we fuse $X_t$ with $D_a$ and $D_v$ to obtain the text-enhanced modality $H^0_t$, 
\begin{equation}
    H^0_t = FC([X_t; D_a; D_v]) \in \mathbb{R}^{T \times d},
\end{equation}
then merge $X_a$ with $D_a$ to obtain the audio-enhanced modality $H^0_a$,
\begin{equation}
    H^0_a = FC([X_a; D_a]) \in \mathbb{R}^{T \times d},
\end{equation}
and we fuse $X_v$ with $D_v$ to obtain the visual-enhanced modality $H^0_v$:
\begin{equation}
    H^0_v = FC([X_v; D_v]) \in \mathbb{R}^{T \times d},
\end{equation}
where FC is fully connected network, [$\cdot$;$\cdot$] represents the concatenation operation.

\subsection{Text-Guided Progressive Fusion}

After obtaining enhanced text, audio, and visual features, we apply Text-Guided Progressive Fusion (TPF) to merge the minor modality. 
TPF consists of $K$ layers of Minor-modality Fusion Units (MFU) and $J$ layers of Core-modality Enhancement Units (CEU), where $K$ and $J$ satisfy the following relationship: $K=J+1$.
The purpose of MFU is to capture text features from different layers of the Transformer as the core modality. 
TPF uses the output of each layer of MFU as a guide and progressively fuses audio and visual features layer by layer, merging them into minor modality features. 
Finally, the core modality and minor modality are further combined to obtain the ultimate fused features.
For ease of discussion, we consider the $k$-th layer of MFU and the $j$-th layer of CEU as one layer of TPF, as illustrated in Figure \ref{fig:tpf}.

\subsubsection{\textbf{Core-modality Enhancement Unit}}

To obtain text features at different levels, we utilize the Transformer Encoder as the $j$-th layer of CEU to extract deeper-level text features:
\begin{equation}
    H^j_t = Trans_j(h^{j-1}_t, {\theta}_{Trans_j}) \in \mathbb{R}^{T \times d},
\end{equation}
where $Trans_j$ and ${\theta}_{Trans_j}$ represent the $j$-th Transformer encoder and corresponding parameters.

\subsubsection{\textbf{Minor-modality Fusion Unit}}

We initialize a minor modality feature $H^0_m$, then use text as a guide to fuse audio and visual features, as shown in Figure \ref{fig:tpf}. The input to MFU includes $H^0_a$, $H^0_v$, the output $Ht$ from the previous CEU layer, and the output $H_m$ from the previous MFU layer. First, we guide the fusion with text for audio, using $H_t$ as $Q$ in Cross-modal Attention, and $H_a$ as $K$ and $V$, performing the fusion with the following formula:
\begin{equation}
  \begin{aligned}
    H_{t \rightarrow a} &= CMA(H^{j-1}_t,H^0_a) \\
    &= softmax(\frac{H^{j-1}_t W_{Q_t} W^T_{K_a} {H^0_a}^T}{\sqrt{d_k}}) W^T_{V_a} {H^0_a}^T \in \mathbb{R}^{T \times d}.
  \end{aligned}
\end{equation}
Similarly, the fusion guided by text for visual features is expressed by the following formula:
\begin{equation}
  \begin{aligned}
    H_{t \rightarrow v} &= CMA(H^{j-1}_t,H^0_v) \\
    &= softmax(\frac{H^{j-1}_t W_{Q_t} W^T_{K_v} {H^0_v}^T}{\sqrt{d_k}}) W^T_{V_v} {H^0_v}^T \in \mathbb{R}^{T \times d}.
  \end{aligned}
\end{equation}
Then we perform a weighted sum of the two fusion vectors, and the minor modality feature is updated by adding it to the sum, serving as the output of this layer of MFU.
\begin{equation}
    H^k_m = H^{j-1}_m + \alpha H_{t \rightarrow a} + \beta H_{t \rightarrow v} \in \mathbb{R}^{T \times d},
\end{equation}
where $H^k_m$ represnets the output minor-modality features of $k$-th MFU, $\alpha$ and $\beta$ are learnable parameters.

\subsubsection{\textbf{Ultimate Multimodal Fusion}}
Finally, the core modality feature $H^J_t$ and the minor modality feature $H^K_m$ are used as the input for the Crossmodal Transformer, resulting in the fused modality vector. Specifically, $H^J_t$ is used as $Q$, and $H^K_m$ is used as $K$ and $V$:
\begin{equation}
  \begin{aligned}
    H &= CMT(H^J_t, H^K_m) \\
    &= softmax(\frac{H^J_t W_{Q_t} W^T_{K_m} {H^K_m}^T}{\sqrt{d_k}}) W^T_{V_m} {H^K_m}^T \in \mathbb{R}^{T \times d}.
  \end{aligned}
\end{equation}

\subsection{Learning Objectives}
Finally, we add a classifier after the Crossmodal Transformer to obtain the final prediction results $\hat{y}$.
For classification tasks, we use the standard cross-entropy loss whereas for regression tasks, we use the mean squared error (MSE) loss as the MSA basic optimization objective, which is:
\begin{equation}
  \begin{aligned}
\mathcal{L} &= -\frac{1}{{N}_{b}}\sum^{{N}_{b}}_{i=0}{y_i} \cdot log{\hat{y}_{i}} \\
  &= \frac{1}{N_b} \sum^{{N}_{b}}_{i=0} | \hat{y}_i - y_i |,
    \end{aligned}
\end{equation}
where $N_b$ is the number of training samples, $\hat{y}$ is the presiction of DEVA, and $y$ is the ground truth.

\section{Experiments}

\subsection{Datasets}
We conduct extensive experiments on three standard multimodal sentiment analysis benchmarks: MOSI~\cite{Zadeh2016MOSI}, MOSEI~\cite{Zadeh2018MOSEI}, and CH-SIMS~\cite{Yu2020CHSIMS}.

\textbf{MOSI}. The MOSI dataset is a popular benchmark dataset in MSA research. This dataset is a collection of YouTube monologues. MOSI contains 2199 subjective words-video clips. These utterances are artificially labeled as consecutive opinion scores between -3 to 3, where -3/+3 represents strong negative/positive sentiment.

\textbf{MOSEI}. The MOSEI dataset is an improvement on MOSI. It contains 23,454 YouTube monologues video segments covering 250 distinct topics from 1,000 distinct speakers. Each utterance also has sentiment consecutive opinion scores between -3 to 3. 

\textbf{CH-SIMS}. The CH-SIMS dataset is a Chinese MSA dataset with fine-grained annotations of modality. The dataset comprises 2,281 video clips collected from various sources, such as different movies and TV serials with spontaneous expressions, various head poses, etc. Human annotators label each sample with a sentiment score from -1 (strongly negative) to 1 (strongly positive).

\begin{table*}[]

\setlength{\tabcolsep}{1mm}{
\begin{tabular}{c cccccc cccccc}
    \toprule
     \multicolumn{1}{c}{\multirow{2}{*}{Methods}} & \multicolumn{6}{c}{MOSI} & \multicolumn{6}{c}{MOSEI} \\
     & Acc-2 & Acc-5 & Acc-7 & F1 & MAE & Corr & Acc-2 & Acc-5 & Acc-7 & F1 & MAE & Corr \\
        \midrule
    TFN* & 77.99/79.08 & - & 34.46 & 77.95/79.11 & 0.947 & 0.673 & 78.50/81.89 & - & 51.6 & 78.96/81.74 & 0.572 & 0.714 \\
    LMF* & 77.90/79.18 & - & 33.82 & 77.80/79.15 & 0.950 & 0.651 & 80.54/83.48 & - & 51.59 & 80.94/83.36 & 0.575 & 0.716 \\
    MulT* & 79.71/80.98 & 42.68 & 36.91 & 79.63/80.95/ & 0.879 & 0.702 & 81.15/84.63 & 54.18 & 52.84 & 81.56/84.52 & 0.559 & 0.733 \\
    ICCN & -/83.07 & - & 39.01 & -/83.02/ & 0.862 & 0.714 & -/84.18 & - & 51.58 & -/84.15 & 0.565 & 0.713 \\
    MISA* & 81.84/83.54 & 47.08 & 41.37 & 81.82/83.58 & 0.776 & 0.778 & 80.67/84.67 & 53.63 & 52.05 & 81.12/84.66 & 0.557 & 0.751 \\
    MAG-BERT & 82.37/84.43 & - & 43.62 & 82.50/84.61 & 0.727 & 0.781 & 82.51/84.82 & - & 52.67 & 82.77/84.71 & 0.543 & 0.755 \\
    PMR & -/82.40 & - & 40.60 & -/82.10 & - & - & -/83.10 & - & 51.80 & -/82.80 & - & - \\
    MFSA & -/83.3 & - & 41.1 & -/83.7 & 0.856 & 0.722 & -/83.8 & - & 53.2 & -/83.6 & 0.574 & 0.724 \\
    FDMER & -/84.6 & - & 44.1 & -/84.7 & 0.724 & 0.788 & -/86.1 & - & 54.1 & -/85.8 & 0.536 & 0.773 \\  
    Self-MM$^\dag$ & 82.54/84.45 & \textbf{52.22} & 45.56 & 82.46/84.44 & \textbf{0.719} & \textbf{0.794} & 82.09/84.76 & 53.54 & 53.65 & 82.43/84.67 & 0.535 & 0.761 \\
    ALMT$^\dag$ & 83.24/85.37 & 50.29 & 44.75 & 83.41/85.46 & 0.738 & 0.776 & 82.34/85.94 & 55.05 & 53.32 & 81.85/85.93 & 0.534 & 0.771 \\
    ConFEDE & 84.17/85.52 & - & 42.27 & 
    84.13/85.52 & 0.742 & 0.784 & 81.65/85.82 & - & \textbf{54.86} & 82.17/85.83 & \textbf{0.522} & \textbf{0.780} \\
        \midrule
    \textbf{DEVA} & \textbf{84.40}/\textbf{86.29} & 51.78 & \textbf{46.32}                          
        & \textbf{84.48}/\textbf{86.30} & 0.730 & 0.787 & \textbf{83.26}/\textbf{86.13} & \textbf{55.32} & 52.26 & \textbf{82.93}/\textbf{86.21} & 0.541 & 0.769 \\
    \bottomrule
\end{tabular}
}
\caption{Comparison on MOSI and MOSEI Datasets.
* represents results obtained from~\cite{Mao2022MSENAAI} and its corresponding GitHub page\textsuperscript{\ref{footnote1}}. Models with $^\dag$ are reproduced under the same conditions. Best results are marked in bold. }
\label{tab:all1}
\end{table*}

\begin{table}
\resizebox{0.48\textwidth}{!}{
  \begin{tabular}{c cccccc}
    \toprule
     Methods & Acc-2 & Acc-3 & Acc-5 & F1 & MAE & Corr \\
    \midrule
    TFN{*} & 78.38 & 65.12 & 39.30 & 78.62 & 0.432 & \textbf{0.591} \\
    LMF{*} & 77.77 & 64.68 & 40.53 & 77.88 & 0.441 & 0.575 \\
    MulT{*} & 78.56 & 64.77 & 37.94 & 79.66 & 0.453 & 0.564 \\
    MISA{*} & 76.54 & - & - & 76.59 & 0.447 & 0.563 \\
    MAG-BERT & 74.44 & - & - & 71.75 & 0.492 & 0.399 \\
    Self-MM$^\dag$ & 77.64 & 64.68 & 41.76 & 77.85 & 0.428 & 0.590 \\
    ALMT$^\dag$ & 78.59 & 64.98 & 40.70 & 78.94 & 0.450 & 0.535 \\
    \midrule
    \textbf{DEVA} & \textbf{79.64} & \textbf{65.42} & \textbf{43.07} & \textbf{80.32} & \textbf{0.424} & 0.583 \\
  \bottomrule
  
\end{tabular}
}
  \caption{Comparison on CH-SIMS. *represents the result is from~\cite{Mao2022MSENAAI} and its corresponding GitHub page \textsuperscript{\ref{footnote1}} . Models with $^\dag$ are reproduced under the same conditions.}
  \label{tab:all2}
\end{table}

\subsection{Evaluation Metrics}
Following the previous works~\cite{Yu2020CHSIMS,Yu2021SelfMM,Zhang2023ALMT}, we present our experimental findings in two distinct formats: classification and regression. In terms of classification, we provide metrics such as Weighted F1 score (F1-Score) and 2-class accuracy (Acc-2). Specifically, for the MOSI and MOSEI datasets, we compute Acc-2 and F1-Score in two configurations: negative / non-negative (including zero) and negative / positive (excluding zero). Moreover, we include additional metrics such as 5-class accuracy (Acc-5) and 7-class accuracy (Acc-7). For the CH-SIMS dataset, we calculate Acc-2, F1, 3-class accuracy (Acc-3), and Acc-5.
Regarding regression, we report Mean Absolute Error (MAE) and Pearson correlation (Corr). In all metrics except MAE, higher values indicate better performance.

\subsection{Baselines}
In order to verify the superiority of our proposed DEVA, we conduct an experimental comparison with the following state-of-the-art baseline models:

\begin{itemize}

\item  Utterance-vector fusion approaches that use tensor-based fusion and low-rank variants: \textbf{TFN} ~\cite{Zadeh2017TFN}, \textbf{LMF} ~\cite{Liu2018LMF}.

\item Models which Learn invariant and specific representations through feature decomposition: \textbf{MISA}~\cite{Hazarika2020MISA}, \textbf{FDMER}~\cite{Yang2022FDMER}, 
\textbf{MFSA}~\cite{Yang2022MFSA},
\textbf{ConFEDE}~\cite{Yang2023ConFEDE}.

\item Models which utilize attention and transformer modules to improve token representations using non-verbal signals: \textbf{MulT}~\cite{Tsai2019MulT}, \textbf{PMR}~\cite{Lv2021ProgressiveMR}, \textbf{ALMT}~\cite{Zhang2023ALMT}.

\item Learning the multimodal and unimodal representations based on the multimodal label and generated unimodal labels: \textbf{Self-MM}~\cite{Yu2021SelfMM}.

\item Learning textbased audio and text-based video features by optimizing canonical loss: \textbf{ICCN}~\cite{Sun2019ICCN}.

\item Model which allows audio and video information to leak into the BERT model for multimodal fusion: \textbf{MAG-BERT}~\cite{Rahman2020MAG}.

\end{itemize}

\begin{table*}[]
\resizebox{1\textwidth}{!}{
\begin{tabular}{l ccccc ccccc ccccc}
    \toprule
    \multicolumn{1}{c}{\multirow{2}{*}{Methods}} & \multicolumn{5}{c}{Modality} & \multicolumn{5}{c}{MOSI} & \multicolumn{5}{c}{CH-SIMS} \\
       & A & V & T & AED & VED & Acc-2 & Acc-7 & F1 & MAE & Corr & Acc-2 & Acc-3 & F1 & MAE & Corr\\ 
        \midrule
    \multirow{3}{*}{w/o ED} & \checkmark & \checkmark & \checkmark &  &  & 82.36/84.15 & 43.00 & 82.46/84.19 & 0.744 & 0.784 & 77.89 & 63.23 & 78.69 & 0.442 & 0.545 \\
      & \checkmark & \checkmark & \checkmark & \checkmark &  & 82.36/84.45 & 40.96 & 82.45/84.48 & 0.749 & 0.781 & 78.11 & 66.30 & 78.03 & 0.425 & 0.577\\
      & \checkmark & \checkmark & \checkmark &  & \checkmark & 83.38/85.21 & 45.77 & 83.43/85.20 & 0.738 & 0.777 & 78.33 & 65.42 & 78.19 & 0.424 & 0.583\\
            \midrule
    w/o AV &  &  & \checkmark & \checkmark & \checkmark & 82.07/84.15 & 43.29 & 82.21/84.21 & 0.751 & 0.773 & \textbf{79.86} & 64.98 & 79.79 & \textbf{0.416} & \textbf{0.604}\\
                \midrule
     \textbf{DEVA} &  \checkmark & \checkmark & \checkmark & \checkmark & \checkmark & \textbf{84.40}/\textbf{86.28} & \textbf{46.21} & \textbf{84.48}/\textbf{86.31} & \textbf{0.737} & \textbf{0.786} & 79.64 & \textbf{65.42}  & \textbf{80.32} & 0.424 & 0.583 \\
    \bottomrule
\end{tabular}
}
\caption{The modality ablation studies on MOSI and CH-SIMS. AED means Audio Emotional Desciption and VED means visual Emotional Description.}
\label{tab:modality ablation}
\end{table*}

\subsection{Comparison of Results}
Tables \ref{tab:all1} and \ref{tab:all2} summarize the comparative results of our proposed method and all baseline models on the MOSI, MOSEI, and CH-SIMS datasets.

As shown in Table \ref{tab:all1}, our proposed DEVA outperforms all baseline models in Acc-2 and F1 (non-negative, negative). On the MOSI dataset, our model exhibits a 0.65\% improvement in Acc-7 over the second-best result. Similarly, on the MOSEI dataset, our model shows a 0.27\% improvement in Acc-5 over the second-best result. In other metrics, our model also approaches state-of-the-art results.

Scenarios in CH-SIMS are more complex than those in MOSI and MOSEI, presenting a greater challenge for multimodal sentiment recognition tasks. However, on the CH-SIMS dataset, our DEVA achieves state-of-the-art performance across all metrics except the Corr indicator. Notably, in binary classification tasks, our method gains a 1.05\% improvement over the best baseline, and it surpasses the best baseline by 1.34\% in the multi-classification metric Acc-5 and outperforming the highest baseline by 0.66\% in F1, indicating the outstanding performance of our model on the challenging CH-SIMS dataset, which features a more complex environment.

Several baseline models also treat text as the core modality, but their results are inferior to those of our model. This demonstrates the effectiveness and advancements of the DEVA model in capturing emotional description.

\footnotetext[1]{\url{https://github.com/thuiar/MMSA/blob/master/results/result-stat.md} \label{footnote1}}

\subsection{Ablation Study and Analysis}

\subsubsection{\textbf{Impacts of Different Modality Combinations}}

In order to investigate the contributions of different modalities, particularly the Audio Emotional Description and Visual Emotional Description, which are carried by the text modality, to the overall performance of the model, we conduct various combinations of modalities on the MOSI and CH-SIMS datasets. We represent Text, Audio, visual, Audio Emotional Description, and Visual Emotional Description as T, A, V, AED, and VED, respectively. The results of the ablation experiments are presented in Table \ref{tab:modality ablation}.

Initially, ablations are performed on Emotional Description (ED), including AED and VED. It can be observed that, compared to the complete model, removing either AED or VED leads to varying degrees of performance degradation across all metrics on the MOSI and CH-SIMS datasets, indicating a positive role of ED in the overall performance of our model. Furthermore, the addition of VED shows a more noticeable improvement in average performance compared to the addition of AED. We infer that the emotional information contained in descriptions of facial expressions and actions might be richer than the emotional information conveyed by broad descriptions of audio signal characteristics such as pitch and volume.

Interestingly, we attempt to replace the original audio and visual source data with modal emotional descriptions for audio and visual modalities, ensuring that the entire model includes only the text modality. The results indicate that, for the MOSI dataset, the model maintains results comparable to traditional MSA inputs. On the CH-SIMS dataset, DEVA (w/o AV) demonstrates competitiveness with our complete model, which simultaneously includes traditional inputs and emotional descriptions. This suggests a novel direction for future MSA research.

\subsubsection{\textbf{Impacts of Different Components}}

\begin{table}
\resizebox{0.48\textwidth}{!}{
  \begin{tabular}{l ccc ccc}
    \toprule
    \multicolumn{1}{c}{\multirow{2}{*}{Methods}} & \multicolumn{3}{c}{MOSI} & \multicolumn{3}{c}{CH-SIMS} \\
     & Acc-5 & MAE & Corr & Acc-2 & Acc-3 & Corr \\

    \midrule
    w/o CEU & 48.98 & 0.751 & 0.780 & 76.36 & 62.36 & 0.551 \\
    w/o MFU & 23.29 & 1.133 & 0.458 & 76.14 & 58.42 & 0.455 \\
    w/o EDG & 48.98 & 0.744 & 0.784 & 77.89 & 63.23 & 0.545 \\
    w/o Fusion Layer & 49.71 & 0.782 & 0.770 & 77.46 & 63.01 & 0.552 \\
    \midrule
     \textbf{DEVA} & \textbf{50.44} & \textbf{0.737} & \textbf{0.786} & \textbf{79.64} & \textbf{65.42} & \textbf{0.583} \\
  \bottomrule
\end{tabular}
}
  \caption{The ablation results by subtracting each component individually on the MOSI and CH-SIMS.}
  \label{tab:components ablation}
\end{table}

To validate the effectiveness of each component in DEVA, we provide the ablation results in Table \ref{tab:components ablation}.
We observe a decrease in all metrics when CEU and MFU are removed (replaced with a single multi-layer Transformer Encoder and addition), with MFU having the greatest impact. 
This demonstrates the significance of the guidance from the text core modality and the progressive interaction between the core modality and the minor modality. Additionally, when removing EDG, i.e., discarding additional emotional description, we find that there is some degradation in model performance, supporting the idea that emotional domain information can enhance the emotion information capture capability in MSA. Finally, when removing the Fusion Layer from the last layer (replaced with a simple concatenation), there is a noticeable decrease in overall scores, highlighting the importance of crossmodal Transformer for effective fusion across different modalities.

\subsection{Fine-grained Study}

\begin{table}
  \resizebox{0.48\textwidth}{!}{
  \begin{tabular}{c cccccc}
    \toprule
     Methods & Polarity & Acc-2 & Acc-4 & Acc-5 & MAE & Corr \\
    \midrule
    \multicolumn{1}{c}{\multirow{6}{*}{DEVA}} & [-3,-2) & {\color{green!100!black}\textbf{56.25}} & {\color{green!100!black}\textbf{30.20}} & {\color{green!100!black}\textbf{8.33}} &{\color{green!100!black}\textbf{1.006}} & 0.065 \\
     & [-2, -1) & {\color{green!100!black}\textbf{55.68
}} & 27.54 & {\color{green!100!black}\textbf{20.76}} & {\color{green!100!black}\textbf{0.677}} & {\color{green!100!black}\textbf{0.128}} \\
     & [-1, 0) & {\color{green!100!black}\textbf{65.51}} & {\color{green!100!black}\textbf{39.65}} & {\color{green!100!black}\textbf{33.62}} & 0.652 & {\color{green!100!black}\textbf{0.196}} \\
     & (0, +1] & {\color{green!100!black}\textbf{59.63
}} & {\color{green!100!black}\textbf{37.61}} & {\color{green!100!black}\textbf{25.85}} & 0.794 & {\color{green!100!black}\textbf{0.140}} \\
     & (+1, +2] & {\color{green!100!black}\textbf{56.66}} & {\color{green!100!black}\textbf{33.33}} & 15.00 & {\color{green!100!black}\textbf{0.674}} & 0.215 \\
     & (+2, +3] & {\color{green!100!black}\textbf{66.66}} & {\color{green!100!black}\textbf{43.75}} & {\color{green!100!black}\textbf{10.41}} & {\color{green!100!black}\textbf{0.760}} & {\color{green!100!black}\textbf{0.316}} \\
    \midrule
    \multicolumn{1}{c}{\multirow{6}{*}{ALMT}} & [-3,-2) & 53.75 & 28.12 & 4.16 & 1.099 & {\color{red!100!black}\textbf{0.102}} \\
     & [-2, -1) & 54.49 & {\color{red!100!black}\textbf{28.74}} & 20.35 & 0.687 & 0.047 \\
     & [-1, 0) & 64.65 & 38.79 & 27.58 & {\color{red!100!black}\textbf{0.584}} & 0.180 \\
     & (0, +1] & 58.71 & 34.86 & 24.77 & {\color{red!100!black}\textbf{0.766}} & 0.130 \\
     & (+1, +2] & 57.50 & 29.16 & {\color{red!100!black}\textbf{17.50}} & 0.693 & {\color{red!100!black}\textbf{0.250}} \\
     & (+2, +3] & 62.50 & 35.41 & 6.25 & 1.033 & 0.312 \\
  \bottomrule
\end{tabular}
}
  \caption{Fine-grained study on MOSI. The {\color{green!100!black}{green number}} indicates the predominance of DEVA, and the {\color{red!90!black}{red number}} indicates the predominance of ALMT.}
  \label{tab:FG}
\end{table}

To explore the performance of DEVA within finer ranges of sentimental polarity, we conduct a fine-grained MSA comparison experiment on the MOSI dataset between DEVA and ALMT. We subdivide the sentimental polarity of MOSI into seven sub-intervals, each with a range of 1. Then, we train models on the entire MOSI dataset and perform regression and classification tasks within each sub-interval separately. The experimental results are shown in Table \ref{tab:FG}. It can be observed that our proposed model performs better than the baseline under the conditions of fine-grained sentimental polarity, especially in binary, ternary, and quinary classification metrics. This demonstrates that our model is more capable of distinguishing subtle changes in sentiment.

\section{Conclusion}

This paper introduces DEVA, a novel method for MSA, which utilizes generated emotional description for progressive fusion. 
DEVA seamlessly integrates the traditional features obtained from pre-trained models with emotional description features, treating text as the core modality and progressively fusing it with audio and visual modalities. 
This approach bridges the gap between different modalities. 
Rigorous experiments on several popular MSA datasets demonstrate the superiority of our proposed method. 
In future work, we aim to enhance the fluency of emotional descriptions to make them more closely daily-life expressions.

\section{Acknowledgments}
This work was supported by the National Natural Science Foundation of China (No. 62302333, 62422210, 62276187) and the Open Research Fund from Guangdong Laboratory of Artificial Intelligence and Digital Economy (SZ) (No. GML-KF-24-16).


\bibliography{aaai25}

\begin{thebibliography}{41}
\providecommand{\natexlab}[1]{#1}

\bibitem[{Aman and Szpakowicz(2007)}]{Aman2007IdentifyingEO}
Aman, S.; and Szpakowicz, S. 2007.
\newblock Identifying expressions of emotion in text.
\newblock In \emph{Proceedings of the 10th International Conference on Text, Speech and Dialogue}, TSD'07, 196–205. Springer-Verlag.
\newblock ISBN 3540746277.

\bibitem[{Bagher~Zadeh et~al.(2018)Bagher~Zadeh, Liang, Poria, Cambria, and Morency}]{Zadeh2018MOSEI}
Bagher~Zadeh, A.; Liang, P.~P.; Poria, S.; Cambria, E.; and Morency, L.-P. 2018.
\newblock Multimodal Language Analysis in the Wild: {CMU}-{MOSEI} Dataset and Interpretable Dynamic Fusion Graph.
\newblock In \emph{Proceedings of the 56th Annual Meeting of the Association for Computational Linguistics (Volume 1: Long Papers)}, 2236--2246.

\bibitem[{Baltru{\v{s}}aitis et~al.(2018)Baltru{\v{s}}aitis, Zadeh, Lim, and Morency}]{Baltruaitis2018OpenFace2F}
Baltru{\v{s}}aitis, T.; Zadeh, A.; Lim, Y.~C.; and Morency, L.-P. 2018.
\newblock OpenFace 2.0: Facial Behavior Analysis Toolkit.
\newblock \emph{2018 13th IEEE International Conference on Automatic Face \& Gesture Recognition (FG 2018)}, 59--66.

\bibitem[{Chen et~al.(2017)Chen, Wang, Liang, Baltru\v{s}aitis, Zadeh, and Morency}]{Chen2017MultimodalSA}
Chen, M.; Wang, S.; Liang, P.~P.; Baltru\v{s}aitis, T.; Zadeh, A.; and Morency, L.-P. 2017.
\newblock Multimodal sentiment analysis with word-level fusion and reinforcement learning.
\newblock In \emph{Proceedings of the 19th ACM International Conference on Multimodal Interaction}, ICMI '17, 163–171.
\newblock ISBN 9781450355438.

\bibitem[{Chen et~al.(2022)Chen, Li, Xu, Wu, Ding, and Zhang}]{Chen2022ViT}
Chen, Z.; Li, B.; Xu, J.; Wu, S.; Ding, S.; and Zhang, W. 2022.
\newblock Towards Practical Certifiable Patch Defense with Vision Transformer.
\newblock \emph{In Proceedings of the 2022 IEEE/CVF Conference on Computer Vision and Pattern Recognition (CVPR)}, 15127--15137.

\bibitem[{Devlin et~al.(2019)Devlin, Chang, Lee, and Toutanova}]{Devlin2019BERTPO}
Devlin, J.; Chang, M.-W.; Lee, K.; and Toutanova, K. 2019.
\newblock {BERT}: Pre-training of Deep Bidirectional Transformers for Language Understanding.
\newblock In \emph{Proceedings of the 2019 Conference of the North {A}merican Chapter of the Association for Computational Linguistics: Human Language Technologies, Volume 1 (Long and Short Papers)}, 4171--4186.

\bibitem[{Doctor et~al.(2016)Doctor, Karyotis, Iqbal, and James}]{Doctor2016AnIF}
Doctor, F.; Karyotis, C.; Iqbal, R.; and James, A.~E. 2016.
\newblock An intelligent framework for emotion aware e-healthcare support systems.
\newblock \emph{2016 IEEE Symposium Series on Computational Intelligence (SSCI)}, 1--8.

\bibitem[{Dong et~al.(2024)Dong, He, Wang, Jin, Ge, Yang, and Jin}]{Dong2024UnveilingID}
Dong, Y.; He, D.; Wang, X.; Jin, Y.; Ge, M.; Yang, C.; and Jin, D. 2024.
\newblock Unveiling Implicit Deceptive Patterns in Multi-Modal Fake News via Neuro-Symbolic Reasoning.
\newblock In \emph{AAAI Conference on Artificial Intelligence}.

\bibitem[{Eyben, W\"{o}llmer, and Schuller(2010)}]{Eyben2010OpensmileTM}
Eyben, F.; W\"{o}llmer, M.; and Schuller, B. 2010.
\newblock Opensmile: the munich versatile and fast open-source audio feature extractor.
\newblock In \emph{Proceedings of the 18th ACM International Conference on Multimedia}, MM '10, 1459–1462. Association for Computing Machinery.
\newblock ISBN 9781605589336.

\bibitem[{Fu et~al.(2021)Fu, Okada, Wang, Guo, Song, Liu, and Dang}]{Fu2021CONSKGCNCS}
Fu, Y.; Okada, S.; Wang, L.; Guo, L.; Song, Y.; Liu, J.; and Dang, J. 2021.
\newblock CONSK-GCN: Conversational Semantic- and Knowledge-Oriented Graph Convolutional Network for Multimodal Emotion Recognition.
\newblock \emph{2021 IEEE International Conference on Multimedia and Expo (ICME)}, 1--6.

\bibitem[{Hazarika, Zimmermann, and Poria(2020)}]{Hazarika2020MISA}
Hazarika, D.; Zimmermann, R.; and Poria, S. 2020.
\newblock MISA: Modality-Invariant and -Specific Representations for Multimodal Sentiment Analysis.
\newblock In \emph{the Proceedings of the 28th ACM International Conference on Multimedia}, 1122–1131.

\bibitem[{Huang et~al.(2020)Huang, Hu, Wang, and Wu}]{Huang2020MultimodalER}
Huang, H.; Hu, Z.; Wang, W.; and Wu, M. 2020.
\newblock Multimodal Emotion Recognition Based on Ensemble Convolutional Neural Network.
\newblock \emph{IEEE Access}, 8: 3265--3271.

\bibitem[{Huang et~al.(2022)Huang, Wang, Chen, Li, Tang, Chu, Chen, Lin, and Ma}]{Huang2021CMUAWatermarkAC}
Huang, H.; Wang, Y.; Chen, Z.; Li, Y.; Tang, Z.; Chu, W.; Chen, J.; Lin, W.; and Ma, K.-K. 2022.
\newblock CMUA-Watermark: A Cross-Model Universal Adversarial Watermark for Combating Deepfakes.
\newblock In \emph{Proceedings of the AAAI Conference on Artificial Intelligence}, volume~36, 989--997.

\bibitem[{Jiang et~al.(2020)Jiang, Li, Hossain, Chen, Alelaiwi, and Al-hammadi}]{Jiang2020ASR}
Jiang, Y.; Li, W.; Hossain, M.~S.; Chen, M.; Alelaiwi, A.; and Al-hammadi, M. 2020.
\newblock A snapshot research and implementation of multimodal information fusion for data-driven emotion recognition.
\newblock \emph{Inf. Fusion}, 53: 209--221.

\bibitem[{Lei, Qian, and Zhao(2016)}]{Lei2016RatingPB}
Lei, X.; Qian, X.; and Zhao, G. 2016.
\newblock Rating Prediction Based on Social Sentiment From Textual Reviews.
\newblock \emph{IEEE Transactions on Multimedia}, 18: 1910--1921.

\bibitem[{Li et~al.(2022)Li, Zhou, Zhang, Liu, Yang, Lian, and Hu}]{Li2022AMOA}
Li, Z.; Zhou, Y.; Zhang, W.; Liu, Y.; Yang, C.; Lian, Z.; and Hu, S. 2022.
\newblock {AMOA}: Global Acoustic Feature Enhanced Modal-Order-Aware Network for Multimodal Sentiment Analysis.
\newblock In \emph{Proceedings of the 29th International Conference on Computational Linguistics}, 7136--7146.

\bibitem[{Liu et~al.(2018)Liu, Shen, Lakshminarasimhan, Liang, Bagher~Zadeh, and Morency}]{Liu2018LMF}
Liu, Z.; Shen, Y.; Lakshminarasimhan, V.~B.; Liang, P.~P.; Bagher~Zadeh, A.; and Morency, L.-P. 2018.
\newblock Efficient Low-rank Multimodal Fusion With Modality-Specific Factors.
\newblock In \emph{Proceedings of the 56th Annual Meeting of the Association for Computational Linguistics (Volume 1: Long Papers)}, 2247--2256.

\bibitem[{Lv et~al.(2021)Lv, Chen, Huang, Duan, and Lin}]{Lv2021ProgressiveMR}
Lv, F.; Chen, X.; Huang, Y.; Duan, L.; and Lin, G. 2021.
\newblock Progressive Modality Reinforcement for Human Multimodal Emotion Recognition from Unaligned Multimodal Sequences.
\newblock \emph{2021 IEEE/CVF Conference on Computer Vision and Pattern Recognition (CVPR)}, 2554--2562.

\bibitem[{Mao et~al.(2022)Mao, Yuan, Xu, Yu, Liu, and Gao}]{Mao2022MSENAAI}
Mao, H.; Yuan, Z.; Xu, H.; Yu, W.; Liu, Y.; and Gao, K. 2022.
\newblock {M}-{SENA}: An Integrated Platform for Multimodal Sentiment Analysis.
\newblock In \emph{Proceedings of the 60th Annual Meeting of the Association for Computational Linguistics: System Demonstrations}, 204--213.

\bibitem[{McFee et~al.(2015)McFee, Raffel, Liang, Ellis, McVicar, Battenberg, and Nieto}]{McFee2015librosaAA}
McFee, B.; Raffel, C.; Liang, D.; Ellis, D. P.~W.; McVicar, M.; Battenberg, E.; and Nieto, O. 2015.
\newblock librosa: Audio and Music Signal Analysis in Python.
\newblock In \emph{Proceedings of the 14th Python in Science Conference 2015 (SciPy 2015)}, 18--24.

\bibitem[{Melville, Gryc, and Lawrence(2009)}]{Melville2009SentimentAO}
Melville, P.; Gryc, W.; and Lawrence, R.~D. 2009.
\newblock Sentiment analysis of blogs by combining lexical knowledge with text classification.
\newblock KDD '09, 1275–1284.
\newblock ISBN 9781605584959.

\bibitem[{Peter and Urban(2012)}]{Peter2012EmotionIH}
Peter, C.; and Urban, B. 2012.
\newblock Emotion in Human-Computer Interaction.
\newblock In \emph{Expanding the Frontiers of Visual Analytics and Visualization}, 239--262.

\bibitem[{Poria et~al.(2020)Poria, Hazarika, Majumder, and Mihalcea}]{Poria2020BeneathTT}
Poria, S.; Hazarika, D.; Majumder, N.; and Mihalcea, R. 2020.
\newblock Beneath the Tip of the Iceberg: Current Challenges and New Directions in Sentiment Analysis Research.
\newblock \emph{IEEE Transactions on Affective Computing}, 14: 108--132.

\bibitem[{Rahman et~al.(2020)Rahman, Hasan, Lee, Bagher~Zadeh, Mao, Morency, and Hoque}]{Rahman2020MAG}
Rahman, W.; Hasan, M.~K.; Lee, S.; Bagher~Zadeh, A.; Mao, C.; Morency, L.-P.; and Hoque, E. 2020.
\newblock Integrating Multimodal Information in Large Pretrained Transformers.
\newblock In \emph{Proceedings of the 58th Annual Meeting of the Association for Computational Linguistics}, 2359--2369. Association for Computational Linguistics.

\bibitem[{Shirian and Guha(2020)}]{Shirian2020CompactGA}
Shirian, A.; and Guha, T. 2020.
\newblock Compact Graph Architecture for Speech Emotion Recognition.
\newblock \emph{ICASSP 2021 - 2021 IEEE International Conference on Acoustics, Speech and Signal Processing (ICASSP)}, 6284--6288.

\bibitem[{Sun et~al.(2020)Sun, Lian, Tao, Liu, and Niu}]{Sun2020MultimodalCD}
Sun, L.; Lian, Z.; Tao, J.; Liu, B.; and Niu, M. 2020.
\newblock Multi-modal Continuous Dimensional Emotion Recognition Using Recurrent Neural Network and Self-Attention Mechanism.
\newblock In \emph{Proceedings of the 1st International on Multimodal Sentiment Analysis in Real-Life Media Challenge and Workshop}, 27–34.

\bibitem[{Sun et~al.(2019)Sun, Sarma, Sethares, and Liang}]{Sun2019ICCN}
Sun, Z.; Sarma, P.~K.; Sethares, W.~A.; and Liang, Y. 2019.
\newblock Learning Relationships between Text, Audio, and Video via Deep Canonical Correlation for Multimodal Language Analysis.
\newblock In \emph{AAAI Conference on Artificial Intelligence}, 8992--8999.

\bibitem[{Tsai et~al.(2019)Tsai, Bai, Liang, Kolter, Morency, and Salakhutdinov}]{Tsai2019MulT}
Tsai, Y.-H.~H.; Bai, S.; Liang, P.~P.; Kolter, J.~Z.; Morency, L.-P.; and Salakhutdinov, R. 2019.
\newblock Multimodal Transformer for Unaligned Multimodal Language Sequences.
\newblock In \emph{Proceedings of the 57th Annual Meeting of the Association for Computational Linguistics}, 6558–6569.

\bibitem[{Vaswani et~al.(2017)Vaswani, Shazeer, Parmar, Uszkoreit, Jones, Gomez, Kaiser, and Polosukhin}]{Vaswani2017AttentionIA}
Vaswani, A.; Shazeer, N.; Parmar, N.; Uszkoreit, J.; Jones, L.; Gomez, A.~N.; Kaiser, L.; and Polosukhin, I. 2017.
\newblock Attention is all you need.
\newblock In \emph{Proceedings of the 31st International Conference on Neural Information Processing Systems}, NIPS'17, 6000–6010. Curran Associates Inc.
\newblock ISBN 9781510860964.

\bibitem[{Wang et~al.(2023)Wang, Dong, Jin, Li, Wang, and Dang}]{Wang2023AugmentingAD}
Wang, X.; Dong, Y.; Jin, D.; Li, Y.; Wang, L.; and Dang, J. 2023.
\newblock Augmenting Affective Dependency Graph via Iterative Incongruity Graph Learning for Sarcasm Detection.
\newblock In \emph{AAAI Conference on Artificial Intelligence}.

\bibitem[{Wu et~al.(2021)Wu, Lin, Zhao, Qin, and Zhu}]{Wu2021TCSP}
Wu, Y.; Lin, Z.; Zhao, Y.; Qin, B.; and Zhu, L.-N. 2021.
\newblock A Text-Centered Shared-Private Framework via Cross-Modal Prediction for Multimodal Sentiment Analysis.
\newblock In \emph{Findings of the Association for Computational Linguistics: ACL-IJCNLP 2021}, 4730--4738.

\bibitem[{Yang et~al.(2022{\natexlab{a}})Yang, Huang, Kuang, Du, and Zhang}]{Yang2022FDMER}
Yang, D.; Huang, S.; Kuang, H.; Du, Y.; and Zhang, L. 2022{\natexlab{a}}.
\newblock Disentangled Representation Learning for Multimodal Emotion Recognition.
\newblock In \emph{Proceedings of the 30th ACM International Conference on Multimedia}, 1642--1651.

\bibitem[{Yang et~al.(2022{\natexlab{b}})Yang, Kuang, Huang, and Zhang}]{Yang2022MFSA}
Yang, D.; Kuang, H.; Huang, S.; and Zhang, L. 2022{\natexlab{b}}.
\newblock Learning Modality-Specific and -Agnostic Representations for Asynchronous Multimodal Language Sequences.
\newblock In \emph{Proceedings of the 30th ACM International Conference on Multimedia}, MM '22, 1708–1717.

\bibitem[{Yang et~al.(2023)Yang, Yu, Niu, Guo, and Xu}]{Yang2023ConFEDE}
Yang, J.; Yu, Y.; Niu, D.; Guo, W.; and Xu, Y. 2023.
\newblock {C}on{FEDE}: Contrastive Feature Decomposition for Multimodal Sentiment Analysis.
\newblock In \emph{Proceedings of the 61st Annual Meeting of the Association for Computational Linguistics (Volume 1: Long Papers)}, 7617--7630.

\bibitem[{Yu et~al.(2020)Yu, Xu, Meng, Zhu, Ma, Wu, Zou, and Yang}]{Yu2020CHSIMS}
Yu, W.; Xu, H.; Meng, F.; Zhu, Y.; Ma, Y.; Wu, J.; Zou, J.; and Yang, K. 2020.
\newblock {CH}-{SIMS}: A {C}hinese Multimodal Sentiment Analysis Dataset with Fine-grained Annotation of Modality.
\newblock In \emph{Proceedings of the 58th Annual Meeting of the Association for Computational Linguistics}, 3718--3727.

\bibitem[{Yu et~al.(2021)Yu, Xu, Yuan, and Wu}]{Yu2021SelfMM}
Yu, W.; Xu, H.; Yuan, Z.; and Wu, J. 2021.
\newblock Learning Modality-Specific Representations with Self-Supervised Multi-Task Learning for Multimodal Sentiment Analysis.
\newblock In \emph{AAAI Conference on Artificial Intelligence}, 10790--10797.

\bibitem[{Zadeh et~al.(2017)Zadeh, Chen, Poria, Cambria, and Morency}]{Zadeh2017TFN}
Zadeh, A.; Chen, M.; Poria, S.; Cambria, E.; and Morency, L.-P. 2017.
\newblock Tensor Fusion Network for Multimodal Sentiment Analysis.
\newblock In \emph{In Proceedings of the 2017 Conference on Empirical Methods in Natural Language Processing}, 1103–1114.

\bibitem[{Zadeh et~al.(2018)Zadeh, Liang, Mazumder, Poria, Cambria, and Morency}]{Zadeh2018MFN}
Zadeh, A.; Liang, P.~P.; Mazumder, N.; Poria, S.; Cambria, E.; and Morency, L.-P. 2018.
\newblock Memory fusion network for multi-view sequential learning.
\newblock In \emph{Proceedings of the AAAI Conference on Artificial Intelligence}, 5634–5641.

\bibitem[{Zadeh et~al.(2016{\natexlab{a}})Zadeh, Zellers, Pincus, and Morency}]{Zadeh2016MOSI}
Zadeh, A.; Zellers, R.; Pincus, E.; and Morency, L.-P. 2016{\natexlab{a}}.
\newblock MOSI: Multimodal Corpus of Sentiment Intensity and Subjectivity Analysis in Online Opinion Videos.
\newblock \emph{IEEE Intelligent Systems}, abs/1606.06259: 82--88.

\bibitem[{Zadeh et~al.(2016{\natexlab{b}})Zadeh, Zellers, Pincus, and Morency}]{Zadeh2016MultimodalSI}
Zadeh, A.; Zellers, R.; Pincus, E.; and Morency, L.-P. 2016{\natexlab{b}}.
\newblock Multimodal Sentiment Intensity Analysis in Videos: Facial Gestures and Verbal Messages.
\newblock \emph{IEEE Intelligent Systems}, 31: 82--88.

\bibitem[{Zhang et~al.(2023)Zhang, Wang, Yin, Liu, Liu, and Yu}]{Zhang2023ALMT}
Zhang, H.; Wang, Y.; Yin, G.; Liu, K.; Liu, Y.; and Yu, T. 2023.
\newblock Learning Language-guided Adaptive Hyper-modality Representation for Multimodal Sentiment Analysis.
\newblock In \emph{Proceedings of the 2023 Conference on Empirical Methods in Natural Language Processing}, 756--767.

\end{thebibliography}

\clearpage
\appendix

\section{Appendix}
\subsection{Hyper-parameters}

\subsubsection{Impacts of Emotional Description Length}

Because facial action units (AUs) are exceptionally diverse, and when assessing a person's emotions, not all facial action units are typically considered, in this section, we investigate the impact of visual Emotional Description (VED) composed of varying numbers of facial action units on the overall model performance.

\begin{figure}[htbp]    
    \centering    
    \begin{subfigure}[b]{\linewidth}    
        \centering  
        \includegraphics[width=\linewidth]{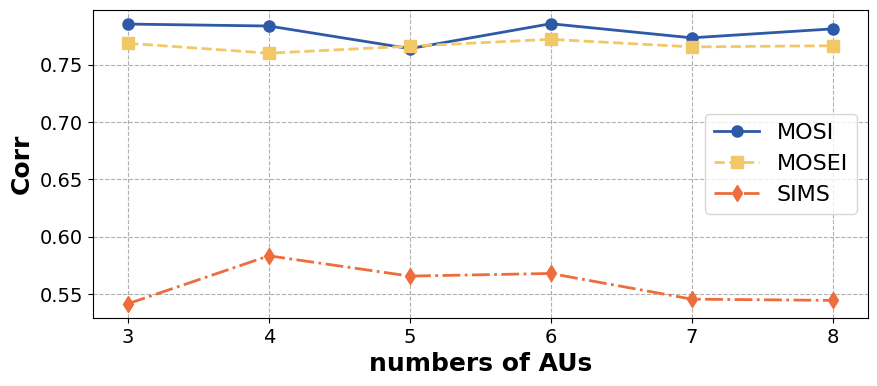}    
        \caption{The impact of the number of action units on correlation.}  
        \label{fig:AUCorr}  
    \end{subfigure}  
    \begin{subfigure}[b]{\linewidth}    
        \centering  
        \includegraphics[width=\linewidth]{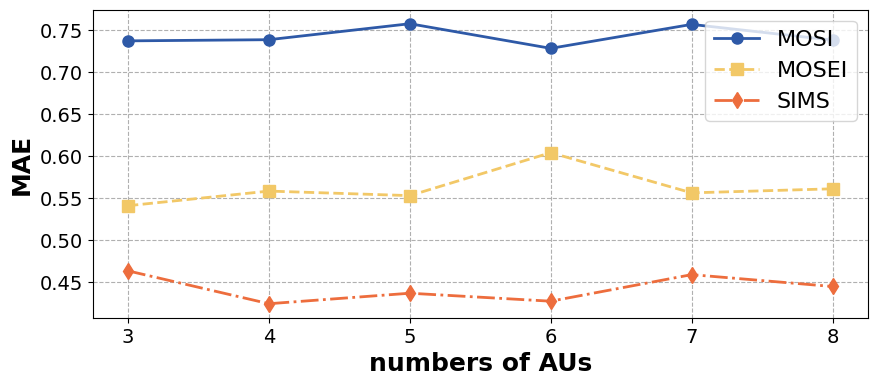}    
        \caption{The impact of the number of action units on MAE.}  
        \label{fig:AUMAE}  
    \end{subfigure}    
    \caption{The impact of the number of Action Units on performance.}   
    \label{fig:fig1}    
\end{figure}

Figures \ref{fig:AUCorr} and \ref{fig:AUMAE} depict the influence of different numbers of AUs on the Corr and MAE metrics in the MOSI, MOSEI, and CH-SIMS datasets, respectively. A greater number of AUs in VED implies a richer description of action units but also introduces redundant information that may have little impact or even adverse effects on emotion judgment. When the number of AUs is excessive, the overall performance is compromised. We observe that the overall performance is optimal when the number of AUs is either 2 or 4.

\subsubsection{Impacts of TPF Depth}

\begin{figure}[ht]
  \centering
  \begin{minipage}[b]{0.49\linewidth}
    \centering
    \includegraphics[width=\linewidth]{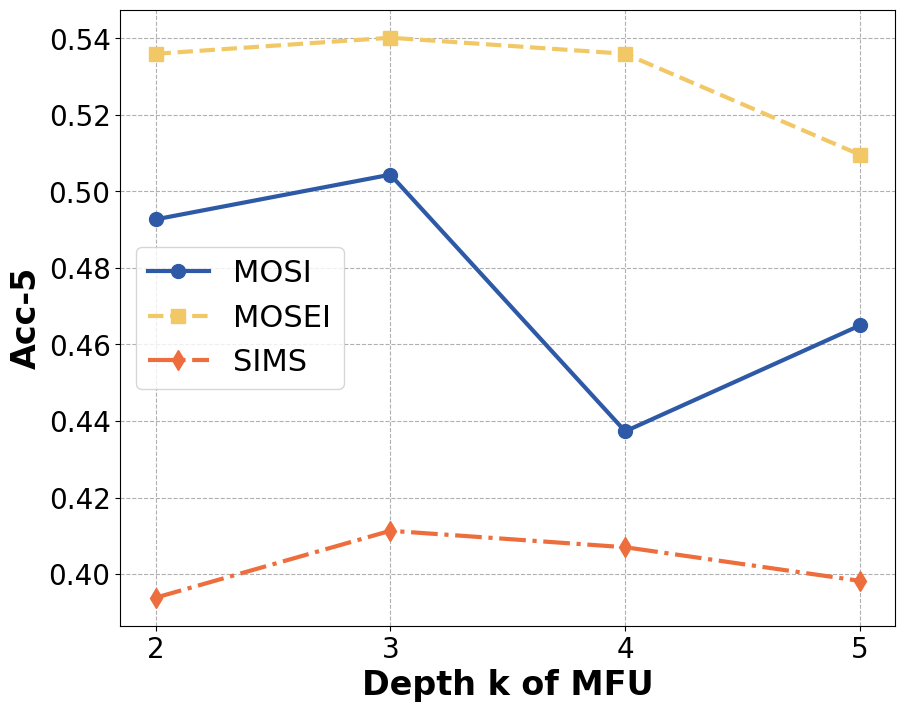}
    (a)
    \label{fig:subfig1}
  \end{minipage}
  \hfill
  \begin{minipage}[b]{0.49\linewidth}
    \centering
    \includegraphics[width=\linewidth]{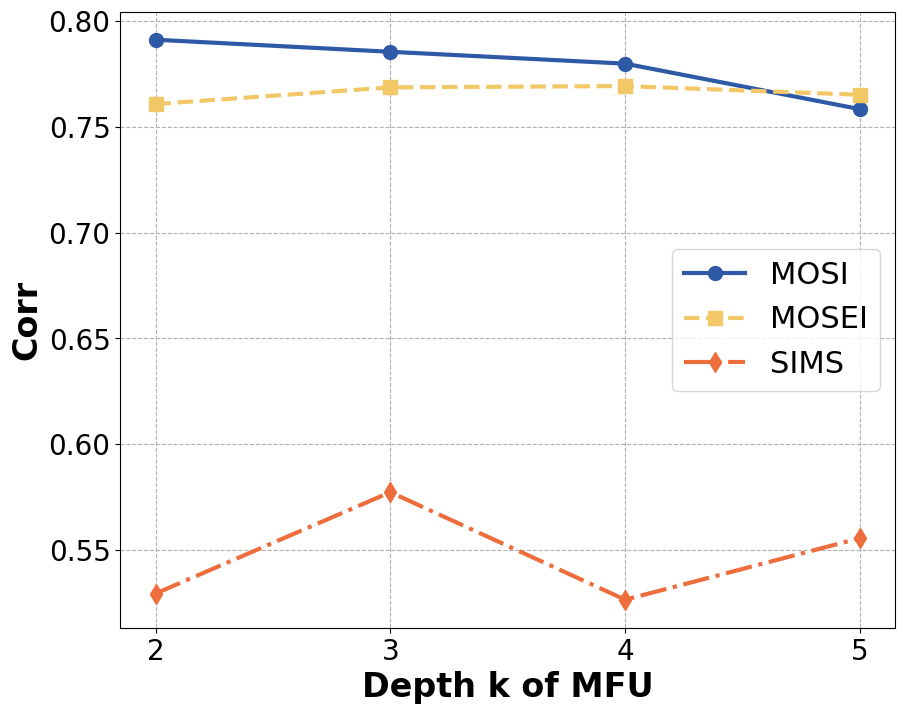}
    (b)
    \label{fig:subfig2}
  \end{minipage}
  \caption{The impact of TPF depth on performance.}
  \label{fig:tpfk}
\end{figure}

Figure \ref{fig:tpfk} illustrates the impact of the number of layers $k$ in TPF on the performance of DEVA across the MOSI, MOSEI, and CH-SIMS datasets. 
It can be observed that when $k$ is set to 3, DEVA achieves optimal performance in terms of Acc-5 and Corr metrics. 
Even when $k$ is set to 2, DEVA exhibits slightly higher Corr on the MOSI and MOSEI datasets compared to when $k$ is set to 3.

\subsubsection{Impacts of T}
\begin{figure}[h]
  \centering  
  \includegraphics[width=\linewidth]{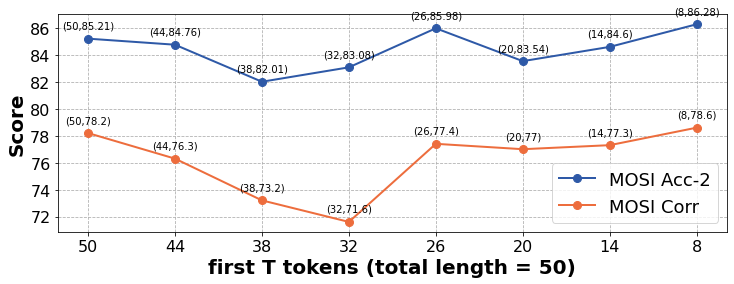}
  \caption{The impact of T.}
  \label{fig:tchoose}
\end{figure}

We follow the settings of ALMT, which suggests that imposing a certain temporal limit ($T$) helps reduce emotion-independent redundant information. Through experiments conducted on the MOSI dataset, we find that the performance reaches its optimal level when $T=8$, as shown in the figure \ref{fig:tchoose}. 

\subsubsection{Impacts of $\alpha$ and $\beta$}
The $\alpha$ and $\beta$ parameters are determined manually through multiple experimental trials. The specific results are shown in Table \ref{tab:ab}. We consider audio and visual modalities to be equally important. Additionally, to restore the dimensionality in Equation 12, we have added a fully connected layer, which also allows the model to learn the importance of audio and visual features.

\begin{table}

  \resizebox{0.48\textwidth}{!}{
  \begin{tabular}{c cccccc}
    \toprule
    hyper & Acc-2 & Acc-5 & Acc-7 & F1 & MAE & Corr \\
    \midrule
    $\alpha$=0.6, $\beta$=1.0 & 85.21 & 48.54 & 41.84 & 85.25 & 0.768 & 0.764 \\
    $\alpha$=0.8, $\beta$=1.0 & 84.91 & \textbf{51.46} & 46.09 & 84.98 & 0.770 & 0.755 \\
    $\alpha$=1.0, $\beta$=0.6 & 84.6 & 50.87 & 45.34 & 84.67 & \textbf{0.733} & 0.782 \\
    $\alpha$=1.0, $\beta$=0.8 & 85.37 & 49.85 & 44.61 & 85.39 & 0.746 & 0.78 \\
    $\alpha$=1.0, $\beta$=1.0 & \textbf{86.28} & 50.44 & \textbf{46.21} & \textbf{86.31} & 0.737 & \textbf{0.786} \\
  \bottomrule
\end{tabular}

}
\caption{$\alpha$ and $\beta$ results on MOSI.}
\label{tab:ab}
\end{table}

\subsection{Different encoding method of VED features}

We apply one-hot encoding (1 means presence of AU and 0 means absence of AU) to the visual emotion description and compare it with our method of mapping to text space, as detailed in Table \ref{tab:onehot}. The experimental results demonstrate that one-hot encoding is significantly inferior to spatial vector encoding.

\begin{table}
  \resizebox{0.48\textwidth}{!}{
  \begin{tabular}{c cccccc}
    \toprule
    Encode & Acc-2 & Acc-5 & Acc-7 & F1 & MAE & Corr \\
    \midrule
    One-hot & 85.37 & 47.67 & 42.57 & 85.48 & 0.739 & 0.780 \\
    BERT & \textbf{86.28} & \textbf{50.44} & \textbf{46.21} & \textbf{86.31} & \textbf{0.737} & \textbf{0.786} \\
  \bottomrule
\end{tabular}
}
  \caption{One-hot encoding and spatial vector encoding results on MOSI.}
    \label{tab:onehot}
\end{table}

\subsection{Different prompt}
In order to verify the effect of the prompt, we test three prompts on the MOSI dataset, as shown in Table \ref{tab:prompt}, and find they have minimal impact on the final outcome.

\begin{table}

  \resizebox{0.48\textwidth}{!}{
  \begin{tabular}{cccccc}
    \toprule
  \multicolumn{6}{c}{ \textbf{Prompt:} The speaker made such an expression:}\\
     Acc-2 & Acc-5 & Acc-7 & F1 & MAE & Corr \\
    \midrule
     84.4/86.28 & 50.44 & 46.21 & 84.48/86.31 & 0.737 & 0.786 \\
    \midrule
   \multicolumn{6}{c}{\textbf{Prompt:} The person formed this expression:}\\
     Acc-2 & Acc-5 & Acc-7 & F1 & MAE & Corr \\
    \midrule
     83.36/85.6 & 49.56 & 44.42 & 83.42/85.7 & 0.738 & 0.786 \\
    \midrule
   \multicolumn{6}{c}{\textbf{Prompt:} The individual wore this facial expression:}\\
     Acc-2 & Acc-5 & Acc-7 & F1 & MAE & Corr \\
    \midrule
     83.67/85.98 & 50.44 & 45.34 & 83.75/85.99 & 0.726 & 0.782 \\
  \bottomrule
\end{tabular}
}
  \caption{Results of using different prompts on MOSI.}
  \label{tab:prompt}
\end{table}

\begin{figure*}[h]
  \centering
  \includegraphics[width=\linewidth]{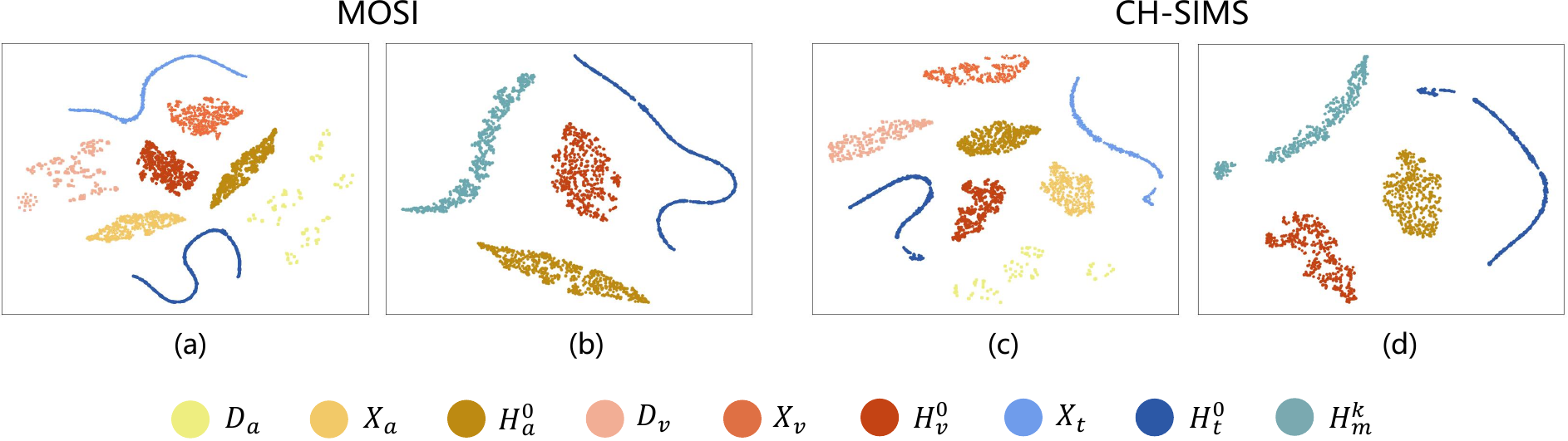}
  \caption{Visualization of Intermediate Features on the MOSI and CH-SIMS Datasets. In the visualization, yellow, red, and blue colors correspond to audio, visual, and text modality-related features, respectively, while cyan indicates minor modality features. The lightest color represents emotional descriptions, whereas the darkest color signifies enhanced features.}
  \label{fig:visual}
\end{figure*}

\subsection{Visualization}

In Figure \ref{fig:visual}, we visualize AED $D_a$, VED $D_v$, preprocessed text representations $X_t$, audio representations $X_a$, visual representations $X_v$, emotional enhanced representations $H^0_t$, $H^0_a$, $H^0_v$, and minor modality representations $H^k_m$ enhanced by MFU on the MOSI and CH-SIMS datasets. It is visualized by t-SNE. (a) and (c) represent emotional descriptions, preprocessed representations, and emotionally enhanced representations on MOSI and CH-SIMS, respectively. It can be observed that emotional enhancement effectively integrates ED with the corresponding modality representations. (b) and (d) show that, even after enhancement, text, audio, and visual modalities still exhibit distribution differences. However, after TPF processing, audio and visual modalities effectively merge into the same distribution, demonstrating that TPF can narrow the distribution gap of minor modalities, enabling more effective fusion with the core modality.

\subsection{Experimental Details}

We used PyTorch to implement our method. The experiments were conducted on a single NVIDIA GeForce RTX 3090. The key parameters are shown in Table \ref{tab:para}.

\begin{table}

    \setlength{\tabcolsep}{1mm}{
  \begin{tabular}{c c c c}
    \toprule
     AU & MOSI & MOSEI & CH-SIMS \\
    \midrule
    Vector Dimension d & 128 & 128 & 128 \\
    Modality Feature Length T & 8 & 8 & 8 \\
    Batch Size & 64 & 64 & 64 \\
    Initial Learning Rate & 1e-4 & 1e-4 & 1e-4 \\
    Optimizer & AdamW & AdamW & AdamW \\
    Epochs & 80 & 80 & 80 \\
    Warm up & \checkmark & \checkmark & \checkmark \\
    dropout & 0.1 & 0.1 & 0.1 \\
  \bottomrule
\end{tabular}
}
  \caption{Hyper-parameters of DEVA.}
  \label{tab:para}
\end{table}

\subsection{Dataset Sizes}

Table \ref{tab:datasets} provides the sizes (number of utterances) in each dataset.
\begin{table}

    \setlength{\tabcolsep}{3.8mm}{
  \begin{tabular}{c cccc}
    \toprule
     Dataset & Train & Valid & Test & Total \\
    \midrule
    MOSI & 1284 & 229 & 686 & 2199 \\
    MOSEI & 16326 & 1871 & 4659 & 22856  \\
    CH-SIMS & 1368 & 456 & 457 & 2281 \\
  \bottomrule
\end{tabular}
}
  \caption{The statistics of MOSI, MOSEI and CH-SIMS.}
  \label{tab:datasets}
\end{table}

\end{document}